\documentclass[10pt,twocolumn,letterpaper]{article}

\usepackage{cvpr}              
\usepackage{comment}
\definecolor{cvprblue}{rgb}{0.21,0.49,0.74}
\usepackage[pagebackref,breaklinks,colorlinks,allcolors=cvprblue]{hyperref}
\usepackage{xcolor}
\definecolor{lime}{RGB}{50, 205, 50}   
\definecolor{orange}{RGB}{255, 140, 0}
\definecolor{violet}{RGB}{128, 0, 128}
\definecolor{latentgreen}{RGB}{124, 252, 0}
\definecolor{latentblue}{RGB}{0, 0, 255}

\def\paperID{*****} 
\def\confName{CVPR}
\def\confYear{2026}

\title{Intra-finger Variability of Diffusion-based Latent Fingerprint Generation}

\author{
 Noor Hussein$^1$ \qquad Anil K. Jain$^1$ \qquad Karthik Nandakumar$^{1,2}$ \\
 $^1$Michigan State University, USA \qquad $^2$MBZUAI, UAE \\
 {\tt\small \{hussei52, jain, nandakum\}@msu.edu}
}

\begin{document}
\maketitle
\begin{abstract}
The primary goal of this work is to systematically evaluate the intra-finger variability of synthetic fingerprints (particularly latent prints) generated using a state-of-the-art diffusion model. Specifically, we focus on enhancing the latent style diversity of the generative model by constructing a comprehensive \textit{latent style bank} curated from seven diverse datasets, which enables the precise synthesis of latent prints with over 40 distinct styles encapsulating different surfaces and processing techniques. We also implement a semi-automated framework to understand the integrity of fingerprint ridges and minutiae in the generated impressions. Our analysis indicates that though the generation process largely preserves the identity, a small number of local inconsistencies (addition and removal of minutiae) are introduced, especially when there are poor quality regions in the reference image. Furthermore, mismatch between the reference image and the chosen style embedding that guides the generation process introduces global inconsistencies in the form of hallucinated ridge patterns. These insights highlight the limitations of existing synthetic fingerprint generators and the need to further improve these models to simultaneously enhance both diversity and identity consistency.
\end{abstract}    
\section{Introduction}
\label{sec:intro}

Recent advancements in generative artificial intelligence (GenAI), such as Generative Adversarial Networks (GANs) and Denoising Diffusion Probabilistic Models (DDPMs), provide the ability to generate large synthetic fingerprint datasets on demand \cite{grosz2024genprint, zhu2023fingerGAN, wyzykowski2023syntheticlatentfingerprintgen, grabovski2024difffinger}. The quality and realism of the generated fingerprint images have also improved dramatically \cite{grosz2024genprint}. Synthetic fingerprint dataset generation typically consists of two-main stages: while the first stage focuses on generating as many unique fingerprint patterns as possible (inter-finger variations), the second stage deals with generating multiple variations of the same ridge pattern (intra-finger variations). 

\begin{figure}[!ht]
\centering
\includegraphics[width=\linewidth]{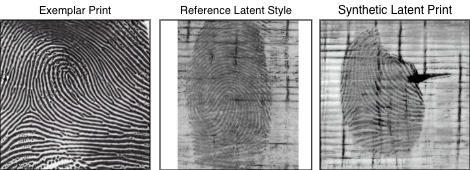} 
\caption{Example of a synthetic latent print. The \textbf{left} image shows the source exemplar print; the \textbf{center} image displays the reference latent style (white tape developed with black wetwop); and the \textbf{right} image shows the resulting synthetic latent print.}
\label{fig:example_lsb}

\end{figure}
In the context of latent fingerprints, the second stage is more critical because exemplar prints are often available in abundance, and the scarcity of mated latent fingerprints is the real bottleneck in both development of Automated Fingerprint Identification Systems and training of latent print examiners (LPE). Furthermore, ground-truth annotation of minutiae in both exemplar and latent prints is also required, especially for LPE training.

\begin{figure*}[h!]
\centering
\includegraphics[width=\linewidth]{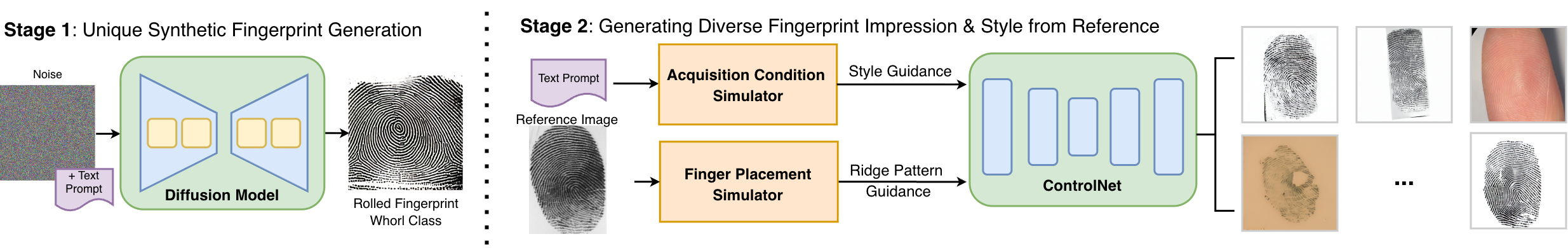} 
\caption{An illustration of core tasks involved in fingerprint synthesis and how they are handled by the GenPrint framework \cite{grosz2024genprint}.}
\label{fig:synthetic_fp_pipeline}
\end{figure*}

The application of existing synthetic fingerprint generators to the latent fingerprint domain is constrained by two key factors. Since current models mainly rely on holistic or random style transfer, they lack the ability to generate fingerprints corresponding to specific forensic scenarios, e.g., ``a latent fingerprint acquired from a glass bottle, lifted with fluorescent powder". In other words, there is a lack of sufficient \textit{diversity} in the latent fingerprints generated by the existing models. Secondly, the stochastic nature of the second-stage generation process can cause subtle alterations or displacements to ridges and minutiae, thereby corrupting the ground-truth identity. Hence, a rigorous effort is needed to quantify the \textit{identity preservation} capability of these generation methods. Both diversity and identity preservation are critical to ensure that intra-finger variations in the generated fingerprints truly mimic real-world variations.


In this work, we focus on the systematic evaluation of intra-finger variability produced by a state-of-the-art diffusion-based synthetic fingerprint generator called GenPrint \cite{grosz2024genprint}. This work makes two contributions.

\begin{itemize}
    \item \textbf{Enhancing the diversity of latent print generation}: The diversity of latent prints generated by GenPrint is limited because it is not possible to precisely specify a latent style. We extend the latent fingerprint generation capability of GenPrint by curating a comprehensive set of over $28,000$ real latent fingerprints from multiple latent datasets that contains latent fingerprints acquired from $40$ surfaces and $15$ latent processing techniques. This latent style bank enables the generation of diverse latent prints depicting different surfaces and processing techniques.
    
    \item \textbf{Quantifying identity consistency of generated fingerprint impressions}: We also introduce a framework for analyzing the integrity of fingerprint impressions generated by GenPrint. Using reference images sampled from a real fingerprint dataset with manually-annotated (ground-truth) minutiae, we generate synthetic fingerprints and verify the validity of minutiae in the generated images in a semi-automated manner. This allows quantification of local inconsistencies (addition and removal of minutiae) introduced by the generator. After aligning the generated impressions with their corresponding reference images, we also identify global ridge patterns that are not present in the reference images but are extrapolated (hallucinated) by the generator. Finally, we also examine the underlying factors that cause these local and global inconsistencies and discuss potential ways to mitigate them.
    
\end{itemize}

\section{Related Work}
\label{sec:related_work}

Synthetic fingerprint generation typically follows a two-stage pipeline as illustrated in \cref{fig:synthetic_fp_pipeline}: (1) generation of unique fingerprint identities (inter-finger variability), and (2) transformation of synthetic identities into diverse styles (intra-finger variability). 

\noindent \textbf{Unique Identity Synthesis}: The primary objective of this stage is to generate a large set of unique finger ridge patterns (identities). Early synthetic fingerprint generators such as SFinGE \cite{cappelli2004sfinge} relied on mathematical models to parametrically simulate ridge flow, density, and orientation field of a fingerprint. However, these approaches fail to generate sufficient number of synthetic identities that can match the real-world variations in the fingerprint ridge patterns.  With the advancement of GenAI, learning of statistical ridge distributions directly from data became possible. Generative Adversarial Networks (GANs) became the standard, with works such as Finger-GAN \cite{minaee2018finger-gan}, PrintsGAN \cite{engelsma2022printsgan}, and others \cite{riazi2020synfi, fahim2020lightweight, bahmani2021high, shoshan2024fpgan} demonstrating the ability to synthesize realistic albeit clean, rolled or plain impressions. More recently, Denoising Diffusion Probabilistic Models (DDPMs) \cite{grabovski2024difffinger} and \cite{grosz2024genprint} utilize latent diffusion to generate unique high-resolution identities, effectively solving the challenge of Stage 1 synthesis to a large extent. 

\noindent \textbf{Fingerprint Style Transfer}: Once a unique identity is established, the challenge shifts to rendering this exemplar print into diverse, valid impressions (e.g., latent, spoof, or sensor specific styles). This resembles a style-transfer problem, where the ridge structure must be preserved while the texture is transformed. In the domain of presentation attack detection, several works have adopted stage 2 in \cref{fig:synthetic_fp_pipeline} to generate synthetic spoofs \cite{abbas2026conditional}. For forensic applications, the goal is to replicate the stochastic degradation of latent prints. Wyzykowski et al. \cite{wyzykowski2023syntheticlatentfingerprintgen} employed a CycleGAN based framework to translate clean rolled fingerprints into latent impressions, mapping inputs into three broad clusters (good, bad, ugly) derived from NIST SD27 \cite{garris2000nist27}. While effective for global degradation, this approach lacks control over specific surfaces. Joshi et al. \cite{joshi2023syntheticLatentStyleTransfer} introduced a neural style transfer pipeline to blend synthetic ridges with real background noise templates. However, these GAN and style-transfer methods often rely on global style injection, which degrades identity consistency. 

\begin{figure}[h]
\centering
\includegraphics[width=\linewidth]{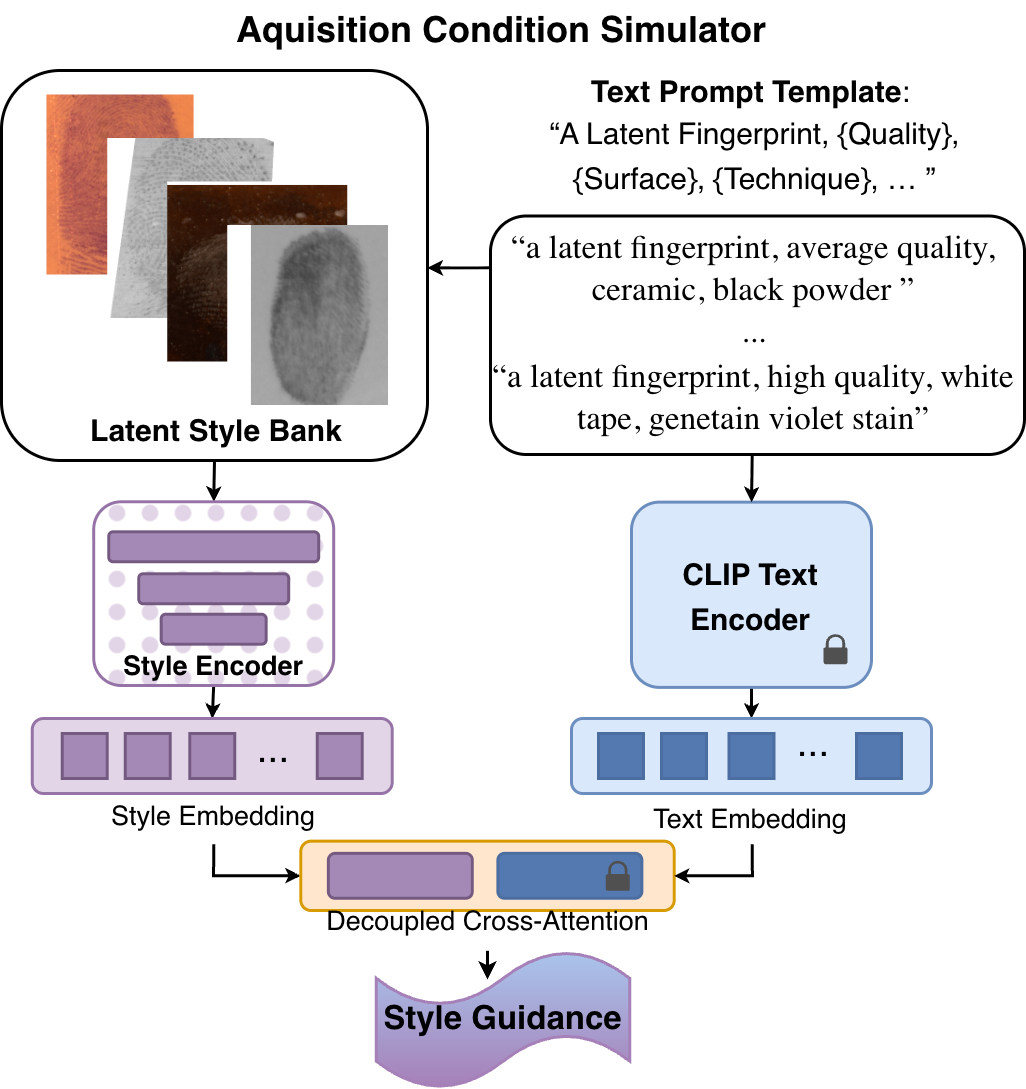} 
\caption{Our proposed modification of GenPrint Stage 2 for diverse latent fingerprint generation with fine-grained style control.}
\label{fig:style_guidance}
\end{figure}

\noindent \textbf{GenPrint}: GenPrint \cite{grosz2024genprint} represents the current state-of-the-art (SOTA) in synthetic fingerprint generation. It integrates both synthesis stages into a single cohesive diffusion-based framework. In the first stage, the model generates a unique synthetic exemplar print from Gaussian noise, conditioned on text prompts that define the class and quality of the fingerprint. In the second stage, GenPrint generates style mates for this identity using linear/nonlinear transformations, masks, SqueezeUNet-based enhancement, and a fine-tuned ControlNet (termed ID-Net). This module freezes the spatial ridge structure of the exemplar print while simulating different finger placement and acquisition conditions via text prompts and style embeddings. Although GenPrint can generate different fingerprint styles (including latent prints), its capability for \textit{latent} fingerprint synthesis is limited by the lack of structured data, resulting in generic or random latent styles. In this work, we focus on \textbf{Stage 2} of the GenPrint pipeline and  construct a \textbf{latent style bank} (see \cref{fig:style_guidance}) that enables precise latent style control. A comparison of our proposed extension against existing latent fingerprint generators is presented in \cref{tab:sota_comparison}. Furthermore, we rigorously analyze the \textbf{identity consistency}, measuring the extent to which the diffusion-based style transfer preserves the minutiae and ridge patterns. 

\begin{table}[!h]
\caption{Comparison between our proposed synthetic latent print generation with existing synthetic latent fingerprint generators.}
\label{tab:sota_comparison}
\centering
\resizebox{\columnwidth}{!}{%
\begin{tabular}{@{}lcccc@{}}
\toprule
\textbf{Method} & \textbf{\shortstack{Source Data}} & \textbf{\shortstack{Surfaces}} & \textbf{\shortstack{Processing\\Techniques}} & \textbf{\shortstack{Latent \\ Style Control}} \\ 
\midrule
Wyzykowski et al. \cite{wyzykowski2023syntheticlatentfingerprintgen}  & 2,288 & NA  & NA & No \\ 
Joshi et al. \cite{joshi2023syntheticLatentStyleTransfer} & 9,000 & \textbf{40} & NA & No \\
GenPrint \cite{grosz2024genprint} & 26,326 & 37 & NA & No \\  
\textbf{Ours} & \textbf{28,357} & \textbf{40} & \textbf{15} & \textbf{Yes} \\
\bottomrule
\end{tabular}%
}
\end{table}

\section{Diversity of Latent Print Generation}
\label{sec:latent_diversity}

\textbf{Vanilla GenPrint Model Pipeline}: In this work, the pre-trained GenPrint model is utilized as follows. The diffusion model in Stage 1 of GenPrint is first used to generate a synthetic exemplar fingerprint image $\mathbf{x}^{s,e}$ using a text prompt with the following template: ``a \textit{rolled} fingerprint image, \{\textit{class}\} pattern, \textit{high} quality, \textit{CrossMatch}'', where the class is one of \{whorl, plain arch, tented arch, left loop, right loop\}. Let $\mathcal{G}$ denote the diffusion (ControlNet) model in Stage 2 of GenPrint that generates a latent fingerprint $\mathbf{x}^{s,v}$ derived from the exemplar print $\mathbf{x}^{s,e}$ using another text prompt with the following template:  ``latent, \{\textit{quality level}\} quality'', where the quality level is one of \{low, average, high\}. Thus, $\mathbf{x}^{s,v} = \mathcal{G}(\mathbf{x}^{s,e},\mathbf{z},\mathbf{t})$, where $\mathbf{z}$ represents a fingerprint style embedding \textit{randomly selected} from a pre-computed style bank and $\mathbf{t}$ is the embedding of the text prompt. The style bank is obtained by applying a style encoder $f$ (a pre-trained VGG-19 network \cite{simonyan2015very}) to each latent fingerprint in the training dataset and the text embedding is obtained by using a CLIP text encoder \cite{radford2021learning}. The style and text embeddings are injected into the diffusion process via decoupled cross-attention layers. 

\subsection{Enhancing GenPrint for Diverse and Controllable Latent Print Synthesis}

Our first goal is to extend the vanilla GenPrint pipeline to generate diverse latent prints with precise latent style control. Here, latent style indicates a combination of the surface from which the latent print is acquired and the technique used for acquisition (henceforth, denoted as \{surface\} + \{technique\}). To achieve this goal, a latent style bank is constructed and the text prompt is extended as:  ``a latent fingerprint, \{\textit{quality level}\} quality, \{\textit{surface}\}, \{\textit{technique}\}''.

\noindent \textbf{Latent Style Bank Construction}: Let $\mathcal{B} = \{(\mathbf{x}_i^{r,e},\mathbf{x}_i^{r,v},\psi_i)\}_{i=1}^{N}$ be a \textbf{real} latent fingerprint dataset, where $\mathbf{x}_i^{r,v}$ denotes the $i^{th}$ latent print, $\mathbf{x}_i^{r,e}$ is the corresponding mated exemplar print, $\psi_i$ represents the latent image style, and $N$ is the number of samples in the dataset. Let $\Psi = \{\psi_m\}_{m=1}^M$ be the collection of $M$ unique latent styles in $\mathcal{B}$. Hence, $\mathcal{B}$ can be partitioned into $M$ subsets $\mathcal{B}_1,\mathcal{B}_2,\cdots,\mathcal{B}_M$, where all latent prints in $\mathcal{B}_m$ have the same style $\psi_m$, $m \in [1,M]$. Let $N_m$ denote the size of the subset $\mathcal{B}_m$. We apply the style encoder $f$ to each latent print $\mathbf{x}_i^{r,v}$ in $\mathcal{B}$ to obtain its style embedding $\mathbf{z}_i = f(\mathbf{x}_i^{r,v})$. Let $\mathcal{Z} = \{\mathbf{z}_i\}_{i=1}^{N}$ be the collection of all style embeddings, which we refer to as the \textbf{latent style bank}. Note that the latent style bank can also be partitioned into $M$ subsets $\mathcal{Z}_1,\mathcal{Z}_2,\cdots,\mathcal{Z}_M$, where each subset contains embeddings of latent prints belonging to a specific style. 

\begin{table}[h!]
\caption{Summary of real latent fingerprint datasets used for constructing the latent style bank.}
\begin{center}
\small 
\vspace{-15pt}
\resizebox{\columnwidth}{!}{%
\begin{tabular}{@{}lcccp{3.7cm}c@{}}
\toprule
\textbf{Dataset} & \textbf{\shortstack{\# Unique \\ Fingers}} & \textbf{\shortstack{\# Latent \\ Images}} & \textbf{Surface} & \textbf{\shortstack{Latent Processing \\ Technique}} & \textbf{\shortstack{In GenPrint's \\ Training Set?}}\\
\midrule
IIITD-SLF \cite{sankaran2012hierarchical-slf} & 120 & 120 & Ceramic & Black powder dusting & Yes \\
IIITD-MUST \cite{malhotra2023must}& 120 & 16327 & 14 surfaces & Black powder, Black wetwop, Cyanoacrylate fuming, DFO, Fluorescent powder, Genetian violet stain, Magnetic black powder, Ninhydrin, White powder, White wetwop & Yes\\
IIITD-MOLF \cite{sankaran2015multisensor-molf} & 1000 & 4400 & Ceramic & Black powder dusting & Yes \\
MSP Latent \cite{yoon2015longitudinal-msp} & 1866 & 2030 & Crime Scene & Powder, Chemical, Ink & Yes\\
NIST-SD27 \cite{garris2000nist27} & 258 & 258 & Crime Scene & NA & No\\
NIST-SD302 \cite{nist302-n2n}& 1019 & 3452 & 22 surfaces & 1,2-Indanedione, Black adhesive-side powder, Black powder, Cyanoacrylate, White adhesive-side powder & Yes\\
LFIW \cite{liu2024latent-lfwild} & 590 & 1770 & Ipad, Wall, Metal & Copper powder & No \\
\midrule
\textbf{TOTAL} & \textbf{4973} & \textbf{28357} & \textbf{40 surfaces} & \textbf{15 techniques} & - \\
\bottomrule
\end{tabular}}
\label{tab:style_bank_dataset_summary}
\vspace{-15pt}
\end{center}
\end{table}

\begin{figure*}[htb]
\centering
\includegraphics[width=\linewidth]{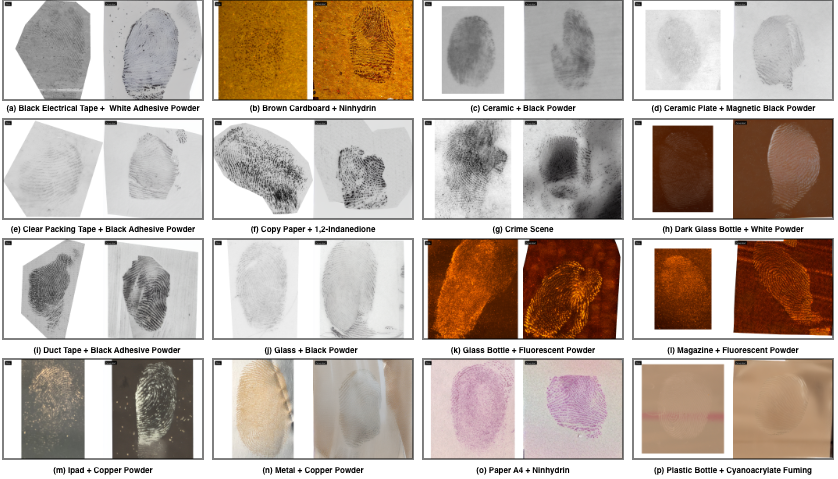} 
\caption{Examples of real latent print (left) and synthetic latent print (right) with different latent styles. Here, the style embedding from the real latent is used for generating the synthetic latent.} 
\label{fig:real_vs_syn}
\end{figure*}

\begin{figure}
    \centering
    \begin{subfigure}[b]{0.25\textwidth}
        \centering
        \includegraphics[width=0.95\textwidth]{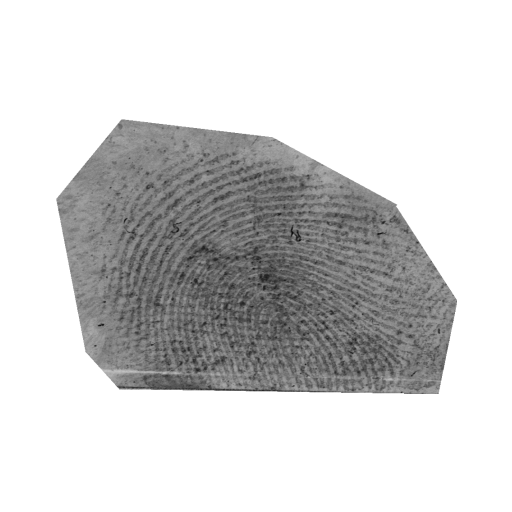}
        \caption{White adhesive-side powder}
        \label{fig:white_adhesive}
    \end{subfigure}
    \hfill
    \begin{subfigure}[b]{0.22\textwidth}
        \centering
        \includegraphics[width=\textwidth]{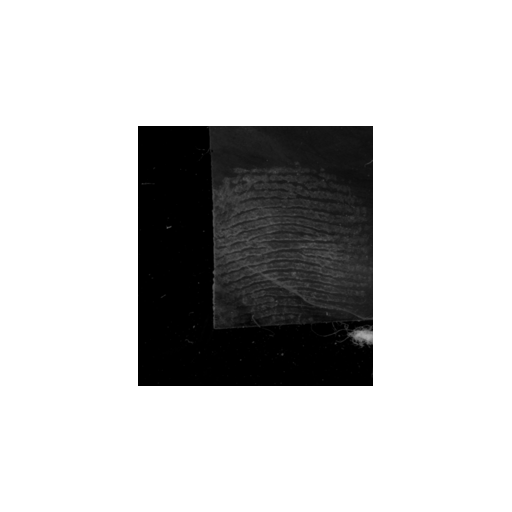}
        \caption{White Wetwop}
        \label{fig:white_wetwop}
    \end{subfigure}
    \caption{Visual comparison of two real latent prints developed on black electrical tape using technically identical methods. Despite the same chemical basis and surface, differences in visual appearance can be noticed.}
    \label{fig:technique_comparison}
\end{figure}

In this work, we construct a diverse latent style bank by curating data from seven real latent fingerprint datasets: IIITD-SLF \cite{sankaran2012hierarchical-slf}, IIITD-MUST \cite{malhotra2023must}, IIITD-MOLF \cite{sankaran2015multisensor-molf}, MSP Latent \cite{yoon2015longitudinal-msp}, NIST-SD27 \cite{garris2000nist27}, NIST-SD302 \cite{nist302-n2n}, and LFWild \cite{liu2024latent-lfwild}. As detailed in  \cref{tab:style_bank_dataset_summary}, this yields a latent dataset of size $N = 28,357$ and provides a source of latent style information covering $40$ unique surfaces and $15$ development techniques. By organizing these datasets into a structured style bank, we extend the GenPrint model to generate latent prints with $M = 45$ latent styles. The full list of $45$ latent styles is shown in \cref{fig:nfiq2_scatter_plot}. This setup allows for \textit{zero-shot style transfer}, enabling synthesis of latent fingerprints on surfaces and techniques it was never explicitly trained on, provided there are latent images with that specific style in the latent style bank.\\
\textbf{Note on Technique Categorization}: Although certain entries in the summary table of the real latent datasets (\cref{tab:style_bank_dataset_summary}), such as ``Wetwop" and ``Adhesive-side powder", represent the same physical development technique, we categorize them separately in this study to account for variations in visual texture (see \cref{fig:technique_comparison}).
Additionally, we note that not only brand formulations but also different application colors (e.g. white powder vs. black powder) may not matter since they are typically standardized to dark-on-light grayscale image by the latent fingerprint examiners. However, we chose to keep them separate to allow more examples of the same group to be synthesized.

\subsection{Validating Latent Print Generation Process}
To validate whether the proposed generation process produces realistic latent prints, we perform the following evaluation. First, we generate a dataset of $100$ synthetic exemplars employing Stage 1 of GenPrint with an equal distribution of five pattern classes. The decision to generate $100$ synthetic exemplars per style is based on the data distribution of our aggregated latent style bank. Real latent datasets vary significantly in size, for instance, the ``glossy magazine + 1,2-indanedione" category contains only 20 images. We selected $N = 100$ as a reference point to ensure statistical fairness and maximize inclusion of categories. Setting a higher threshold would force the exclusion of rare but critical style types from the one-to-one matching analysis. Consequently, \cref{tab:megamatcher_results} only includes categories that met this $100$ image minimum to ensure consistency.
For each synthetic exemplar, we then generate $M=45$ different latent prints with each latent print belonging to a different style. Some examples of synthetic latent prints generated with different latent styles are shown in \cref{fig:real_vs_syn}. To quantify the utility of the generated latent prints, we evaluate the recognition performance based on the synthetic images as well as the quality distribution of the synthetic latent prints. 

\begin{figure*}[ht]
\centering
\includegraphics[width=\linewidth]{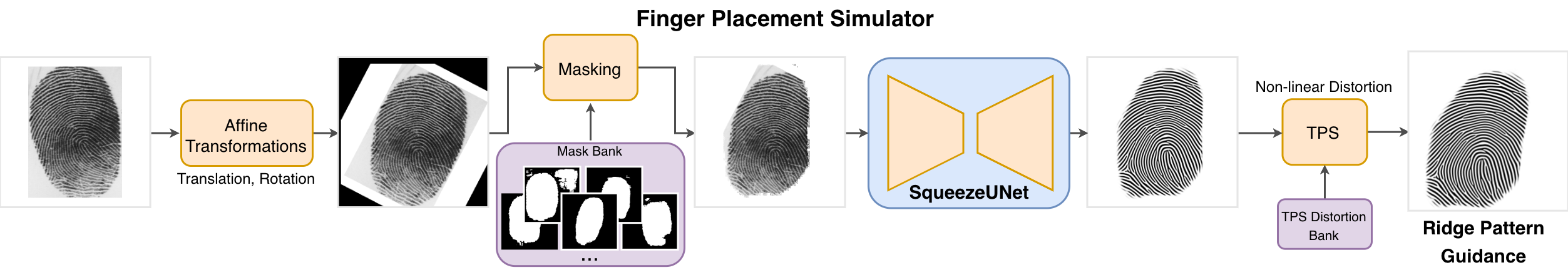} 
\caption{Detailed overview of the finger placement simulator that provides ridge pattern guidance.}
\label{fig:ridge_pattern_guidance}
\vspace{0.1cm}
\end{figure*}

\begin{figure}[ht]
    \centering
    \includegraphics[width=\linewidth]{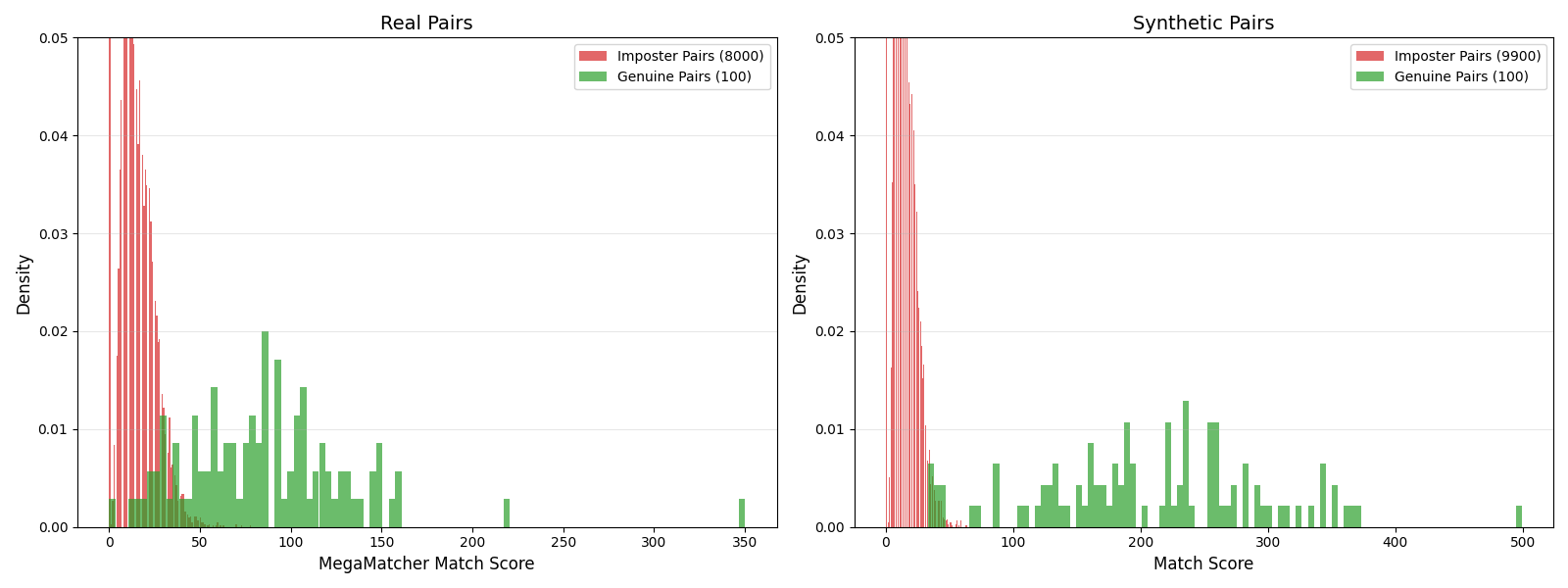}
    \caption{Genuine and impostor match score distributions for real and synthetic pairs. In brackets is the number of matching pairs.}
    \label{fig:match_score_distributions}
\end{figure}

\noindent \textbf{Recognition Performance}: To verify that the style transfer process does not degrade the ridge structure and preserves finger identity, we match the synthetic latent prints with synthetic exemplar prints using MegaMatcher SDK v2025.1.0 \footnote{\label{megamatcher}https://www.neurotechnology.com/megamatcher.html}. A higher match score between a latent print and its corresponding mated exemplar is an indication that the style transfer is purely textural and has not destroyed the identity. Note that this experiment is performed individually for each latent style. We evaluate verification performance using the True Match Rate (TMR) at a fixed threshold score of $48$ (according to MegaMatcher's SDK documentation\textsuperscript{\ref{megamatcher}}, this threshold approximately corresponds to a False Match Rate (FMR) of $0.01$\%) across different latent styles. We refer to this matching protocol as \textit{Synthetic Pair} in \cref{tab:megamatcher_results}. As a baseline for evaluating this verification performance, we also consider two other matching protocols. First, we match the real latent prints in the curated dataset $\mathcal{B}$ with their corresponding real exemplars and denote this setting as \textit{Real Pair}. Next, we sample a subset of real exemplars from $\mathcal{B}$ and use them as reference (completely bypassing Stage 1 of GenPrint) to generate synthetic latent prints. Matching between these synthetic latent prints and their corresponding real exemplars is referred to as \textit{Hybrid Pair}. The results of all the above three matching protocols for a subset of latent styles are summarized in \cref{tab:megamatcher_results}. These results demonstrate that the match scores between the synthetic pairs are comparable to those of the real mated pairs. The genuine (mated) and impostor (non-mated) match score distributions for the generated and real pairs are also quite similar as shown in \cref{fig:match_score_distributions}, further demonstrating the preservation of intra- and inter-finger variations during synthetic fingerprint generation. This confirms that the style transfer process mostly preserves the global ridge flow and consequently, the synthetic latent prints maintain a high similarity with their mated exemplar prints. 

\begin{table}[t]
\caption{Verification performance for different latent styles (surface+technique combination) based on MegaMatcher. Here, real pair indicates matching real exemplar and latent prints, synthetic pair indicates matching synthetic exemplar and latent prints, and hybrid pair indicates matching real exemplar and synthetic latents.}
\centering
\resizebox{\columnwidth}{!}{%
    \begin{tabular}{@{}llccc@{}}
    
    \toprule
     &  &\multicolumn{3}{c}{\textbf{ TMR (\%) @ FMR=0.01\%}} \\
    \cmidrule(lr){3-5}
    \textbf{Surface} & \textbf{Technique} & \textbf{\textit{\shortstack{\quad Real \quad \quad \\ Pair \quad }}} & \textbf{\textit{\shortstack{Synthetic \\ Pair}}} & \textbf{\textit{\shortstack{Hybrid \quad \\ Pair \quad }}}\\
    \midrule
    Apple iPhone 5s & Black powder  & 79 & 93 & 99\\
    Black Dustbin Bag & Cyanoacrylate fuming & 79  & 95 & 90\\
    Ceramic & Black powder & 75 & 95 & 96\\
    Ceramic Plate & Black powder & 86 & 96 & 96\\
    Ceramic Plate & Magnetic black powder & 89 & 95 &  97\\
    Clear Tape & Black wetwop & 89 & 95 & 92\\
    Crime Scene & Unknown technique  & 84 & 96 & 99\\
    Cylindrical Tube & Black powder & 83 & 94 & 98\\
    Duct Tape & Black adhesive-side powder & 80 & 99  & 98\\
    Glass & Black powder & 85 & 91 & 98\\
    Paper A4 Sheet & DFO & 76 & 91 & 93\\
     Plastic Container & Black powder & 78 & 95 & 89\\
    White Tape & Black wetwop & 97 & 92 & 96\\
    Ziplock Bag & Cyanoacrylate fuming & 88 & 93 & 91\\
    \bottomrule
    \end{tabular}
}
\vspace{-0.3cm}
\label{tab:megamatcher_results}
\end{table}

\begin{figure*}[h]
    \centering
    \includegraphics[width=\linewidth]{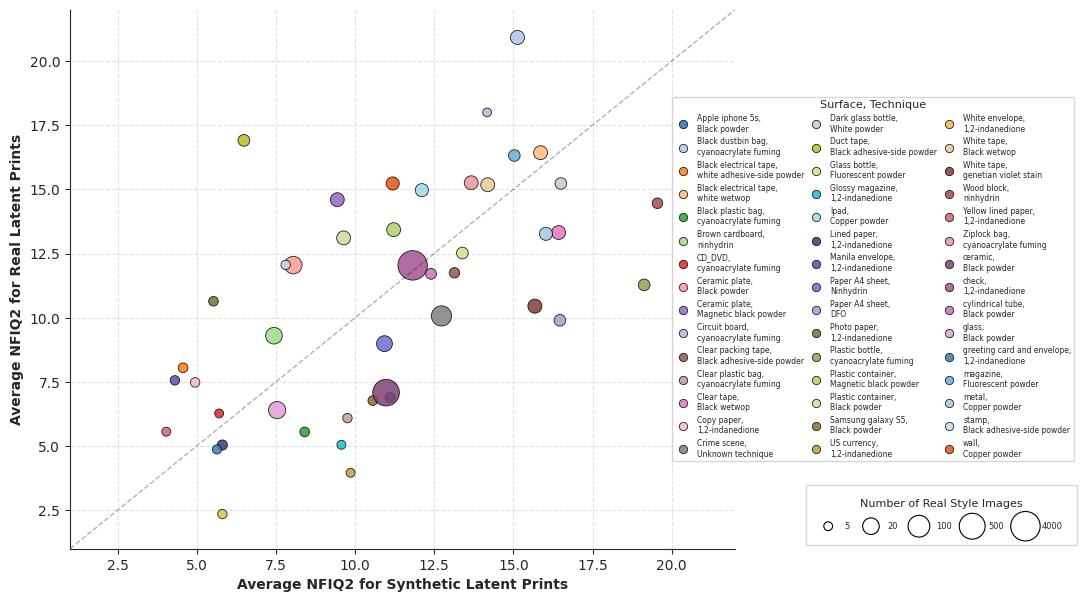}
    \caption{The average NFIQ2 quality for synthetic vs. real latent prints. The circle size is proportional to the size of the latent style bank for that style. The dashed line represents the case where the average quality values of real and synthetic latent prints are equal.}
    \label{fig:nfiq2_scatter_plot}
\end{figure*}

\begin{figure}[h]
    \centering
    \includegraphics[width=\linewidth]{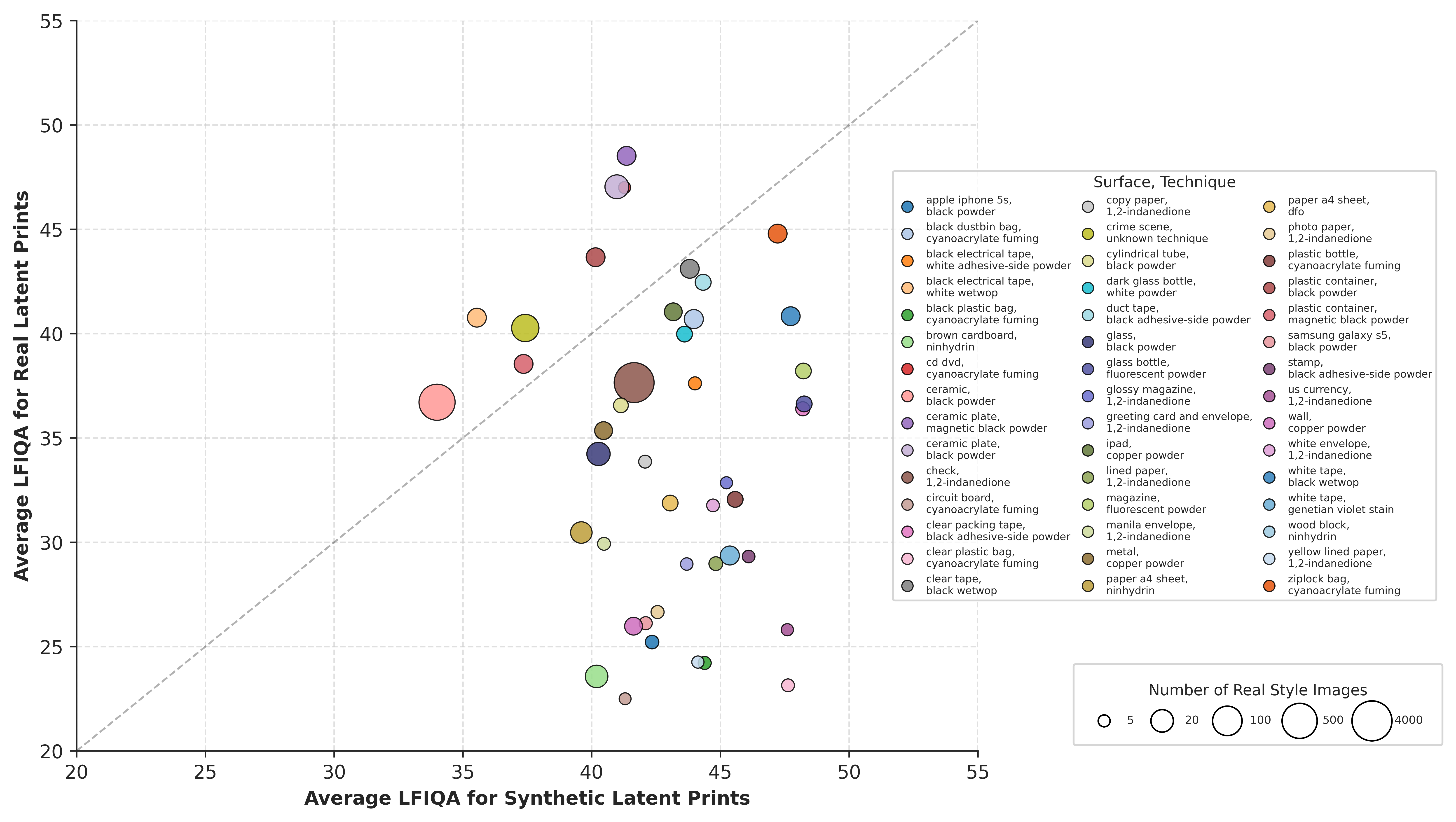}
    \caption{Scatter plot correlating the average LFIQA scores of synthetic latent against real latent prints from the same styles.}
    \label{fig:lfiqa_scatter_plot}
\end{figure}

\noindent \textbf{Quality Analysis}:
An ideal latent print generator must not only replicate the look of a reference style, but also preserve the relative image quality distributions. For example, latent prints acquired from a cardboard surface are typically of lower quality than those acquired from a ceramic surface. If the generator produces high-quality prints for every surface, it does not reflect the forensic reality. Hence, synthetic latent prints of different styles should have a quality distribution similar to their real counterparts of the same style. We used NFIQ2 \cite{tabassi2021nfiq} quality score to measure this alignment. NFIQ2 assigns integer quality scores in range $[0,100]$ based on image clarity, ridge flow, and quality of the minutiae. We compare the average quality scores of each real latent style with the generated synthetic latents. \cref{fig:nfiq2_scatter_plot} shows a scatter plot between the average quality of each real and synthetic latent prints for all $45$ latent styles. The size of the circle in the plot is proportional to $N_m$, which is the number of samples in the style bank for style $\psi_m$. Since the data points are close to the diagonal line, we can conclude that the generative pipeline successfully preserves the intrinsic quality profile of each latent style. If a specific surface naturally yields low quality latent prints (e.g., ``Yellow lined paper + 1,2-indanedione" towards the lower-left), the generative model accurately reproduces that same low quality fingerprint on average rather than artificially inflating the quality. However, since NFIQ2 was designed primarily for exemplar (plain/rolled) fingerprints, we further validate our results using Latent Fingerprint Image Quality Assessment (LFIQA) \cite{huang2026fiqa}. Unlike general quality metrics, LFIQA is specifically trained to predict the utility of a latent print for automated search engines. As shown in \cref{fig:lfiqa_scatter_plot}, the average LFIQA scores for synthetic latents correlate with those of real latents across different latent styles.

\begin{figure}[!htb] 
    \centering
    \begin{minipage}{\columnwidth}
        \centering
        \includegraphics[width=0.8\linewidth]{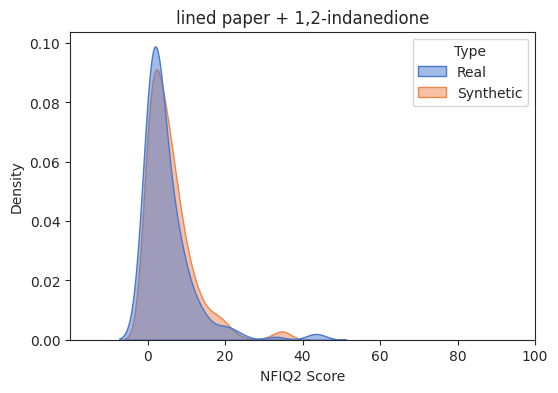}
        \subcaption{Lined Paper}
        \label{fig:lined_paper}
    \end{minipage}
    \hfill
    \begin{minipage}{\columnwidth}
        \centering
        \includegraphics[width=0.8\linewidth]{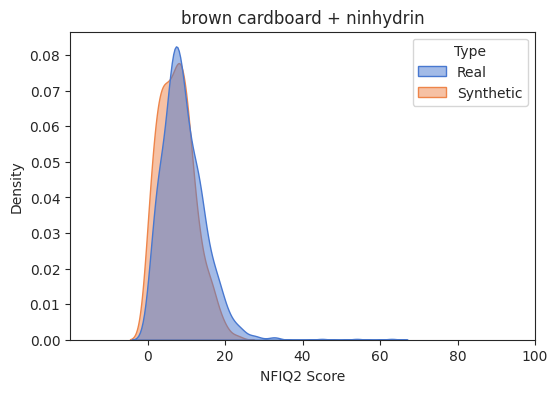}
        \subcaption{Brown Cardboard}
        \label{fig:brown_cardboard}
    \end{minipage}
    
    \caption{Examples of NFIQ2 quality score distributions for real and synthetic latent prints corresponding to two styles.}
    \label{fig:comparison_combined}
\end{figure}

Furthermore, we compare the quality score distributions between real and synthetic latent prints for each style, and two examples are shown in \cref{fig:lined_paper} and \cref{fig:brown_cardboard}. We observe that synthetic latent prints not only mimic the target style (see \cref{fig:real_vs_syn}), but also replicates the quality distribution of real data, as shown in \cref{fig:comparison_combined}. For both latent styles (lined paper+1,2-Indanedione and brown cardboard+ninhydrin), the NFIQ2 quality histograms between real and synthetic latent prints largely overlap. This confirms that the style guidance successfully captures the characteristics of each surface and the generation process effectively produces the expected diversity in latent print generation.

\section{Quantifying Identity Consistency}
\label{sec:identity_consistency}

One of the critical aspects of intra-finger variability in synthetic fingerprint generation is the need to ensure that variations generated from the exemplar print faithfully follow the ridge structure and minutiae details of the exemplar print. Ideally, the transformation from an exemplar print to any style should change only acquisition conditions like contrast, noise, texture (see \cref{fig:style_guidance}), and also introduce finger placement variations in terms of affine transformations and non-linear distortion (see \cref{fig:ridge_pattern_guidance}). Since diffusion models are inherently stochastic, there is a significant risk of ``hallucination'' \cite{aithal2024understandinghallucinations}, where the generative process might inadvertently alter, remove, or create new minutiae and ridges. 


Benchmarking the recognition performance of synthetic fingerprints via an automated matcher serves as a good initial check on identity consistency. However, this evaluation is insufficient because automated fingerprint matching is inherently designed to be robust against intra-finger variations such as non-linear distortions, spurious minutiae, and partial feature loss \cite{maltoni2009handbook}. Consequently, even a high match score from an automated matcher does not indicate an exact match. In fact, a high match score between a synthetic fingerprint impression and its mated exemplar can be observed even though the generated impression contains subtle hallucinations including added/displaced ridges and spurious/missing minutiae. Therefore, the only way to precisely quantify the identity consistency is through manual or semi-automated annotation. We now describe a semi-automated framework for quantifying identity consistency of synthetic fingerprints generated using GenPrint.

\subsection{Data Preparation}

Our goal is to accurately evaluate the ability of the diffusion model in Stage 2 of GenPrint to preserve minutiae and ridge information when it generates a new impression based on a given exemplar print. Ideally, this evaluation must be conducted based on the synthetic exemplar-latent pairs generated in Section \ref{sec:latent_diversity}. However, both the synthetic exemplar and latent prints generated earlier do not have any ground-truth minutiae annotations. Furthermore, since the latent prints also have lower image quality on average, it is hard to manually annotate sufficient number of prints. Hence, to simplify the annotation process, we use the existing IBM-HURSLEY \cite{prabhakar2000ibm-hursley} database, which has $900$ slap fingerprints with manually annotated ground-truth minutiae by fingerprint examiners. We select a subset of $100$ fingerprints from this dataset as real exemplar prints. For each of these $100$ real exemplars, we generate a synthetic impression using the following text prompt: ``A slap fingerprint image, high quality, CrossMatch, FTIR Optical''. We used this prompt to synthesize fingerprints with a fixed high-quality target style so that any structural errors can be solely attributed to the generative process rather than style variations.

To assess the behavior of the generative process under changes in image quality of the real exemplar prints, the $100$ selected exemplars consist of $50$ high quality and $50$ low quality images. This quality categorization is achieved by obtaining the NFIQ2 quality scores ($q$) for all images in the IBM-HURSLEY dataset and dividing the images into three quality bins as follows:

\begin{align*}
     \text{Quality Class =}\left\{ \begin{array}{lcl}
\text{high} & \mbox{if} & q > \mu + \sigma \\
\text{average} & \mbox{if} & \mu - \sigma \leq q \leq \mu + \sigma  \\
\text{low} & \mbox{if} & q < \mu - \sigma \\
\end{array}\right.
\end{align*}

\noindent Here, $\mu$ and $\sigma$ denote the mean and standard deviation of quality scores based on a large aggregated dataset of real slap fingerprints \cite{grosz2024genprint}. In our implementation, these statistics are estimated as $\mu = 47.67$ and $\sigma = 23.16$. 

In summary, we create a dataset of $100$ mated slap fingerprint pairs $\{(\mathbf{x}^{r,e}_i,\mathbf{x}^{s,u}_i,)\}_{i=1}^{100}$, where the exemplar slap print $\mathbf{x}^{r,e}_i$ is sampled from the IBM-HURSLEY dataset and its mated slap impression $\mathbf{x}^{s,u}_i$ is synthetically generated using Stage 2 of GenPrint. Let $\mathbf{M}^{gt}_i$ denote the (manually annotated) ground-truth minutiae set corresponding to the real exemplar $\mathbf{x}^{r,e}_i$. We also compute the binary foreground mask of the fingerprint region in $\mathbf{x}^{r,e}_i$ using the Verifinger SDK v12.4\footnote{\label{verifinger12.4}https://www.neurotechnology.com/verifinger.html} and denote this foreground mask as $\mathbf{B}^{gt}_i$. Unlike latent prints, the real exemplar prints have very less background noise. Hence, the extracted foreground masks can be considered as ground-truth information.   
As shown in \cref{fig:ridge_pattern_guidance}, Stage 2 of GenPrint employs the following four steps to simulate finger placement variations. (1) A randomly sampled affine transformation is applied to the exemplar print. (2) A fingerprint region mask is randomly selected from a mask bank (pre-computed from the large aggregated dataset of real slap fingerprints) and applied to the transformed print to determine the foreground shape of the generated synthetic print. This step simulates partial fingerprint presentation by the user. (3) The ridge information of the masked fingerprint is enhanced using a SqueezeUNet module \cite{beheshti2020squeeze}. (4) A non-linear distortion map represented as thin plate splines (TPS) is randomly selected from a pre-computed TPS bank and is applied to distort the enhanced ridges. The final ridge map obtained at the end of these four steps is used by the ControlNet as ridge pattern guidance. Since the exact transformations applied to the exemplar print $\mathbf{x}^{r,e}_i$ are known, we can apply the same transformations to the ground-truth minutiae set $\mathbf{M}^{gt}_i$ and the ground-truth mask $\mathbf{B}^{gt}_i$ to obtain the expected minutiae set $\mathbf{M}^{exp}_i$ and the expected mask $\mathbf{B}^{exp}_i$, respectively. If there is perfect identity preservation during synthetic print generation, one would expect the synthetic print $\mathbf{x}^{s,u}_i$ to have exactly the same minutiae available in $\mathbf{M}^{exp}_i$ and the foreground mask of the synthetic print must be identical to $\mathbf{B}^{exp}_i$. Any deviation from this expected behavior can be considered as hallucination (errors) introduced by the generation process.

\begin{figure*}[h!]
\centering
\includegraphics[width=\linewidth]{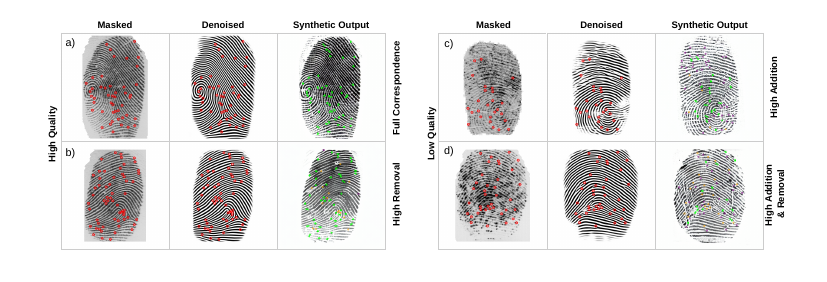} 
\caption{Local error analysis. The first column displays the ground-truth minutiae in \textcolor{red}{red} overlaid on the real exemplar image from the IBM-HURSLEY dataset. The second column shows the expected minutiae in \textcolor{red}{red} overlaid on the ridge pattern guidance that is fed as input to the ControlNet module. These minutiae are expected to be present in the synthetic print generated by ControlNet. In the third column, the markers illustrate the manual annotation results on the synthetic print: \textcolor{lime}{green} indicates a \textit{\textbf{matched}} minutia between the exemplar and synthetic prints, \textcolor{orange}{orange} indicates \textit{\textbf{missing}} minutia that is removed by the generator, and \textcolor{violet}{purple} highlights \textit{\textbf{spurious}} minutiae introduced during generation.}
\label{fig:example_manual_annotation}
\end{figure*}

\subsection{Quantification of Local and Global Errors}

\textbf{Local Errors}: For each synthetic print $\mathbf{x}^{s,u}_i$, we apply the Verifinger SDK v12.4 to extract the minutia set $\mathbf{M}^{gen}_i$ and the binary foreground mask $\mathbf{B}^{gen}_i$. Since there may be errors in the automated minutiae extraction process, we cannot directly compare $\mathbf{M}^{gen}_i$ with $\mathbf{M}^{exp}_i$ to identify the local minutiae errors. Hence, we developed a custom manual annotation tool to facilitate a precise comparison between the expected and generated minutiae sets. Using this tool, we overlay $\mathbf{M}^{gen}_i$ against $\mathbf{M}^{exp}_i$ on the generated image. Next, we categorize each minutia point in the two sets into one of the following three cases: (i) \textit{\textbf{matched}} (present in both the real exemplar and the generated synthetic prints), (ii) \textit{\textbf{missing}} (present in the real exemplar, but not present in the synthetic print), or (iii) \textit{\textbf{spurious}} (not present in the real exemplar, but present in the synthetic print).  


A minutia pair (one from the real exemplar and the other from the synthetic print) is considered as \textit{\textbf{matched}}, if they are within a tolerance bounding box of size $9$ pixels. This tolerance level is determined by the average width of a ridge in the aggregated dataset and it accounts for the minor variations in the minutia localization processes (of both human examiners and automated algorithms). Note that some of the spurious and missing minutiae may be due to errors introduced by the automated extraction algorithm. For instance, a minutia point may be present in the generated image but not extracted by the algorithm. In this case, we add the minutia point using the custom tool and consider it as \textit{\textbf{matched}} if it is also present in the exemplar print. Similarly, when a spurious minutia point is introduced by the extractor without any visual evidence in the generated image, we use the custom tool to remove this minutia point. Examples of this manual annotation step are shown in \cref{fig:example_manual_annotation}.

After applying all the above manual post-processing steps, we count the number of ``true'' \textit{\textbf{missing}} and \textit{\textbf{spurious}} minutiae, which are solely introduced due to the stochasticity of the synthetic image generation process. For each real exemplar-synthetic print pair $(\mathbf{x}^{r,e}_i,\mathbf{x}^{s,u}_i)$ in the dataset, we find the number of \textit{\textbf{matched}}, \textit{\textbf{missing}}, and \textit{\textbf{spurious}} minutiae as $\alpha_i$, $\beta_i$, and $\gamma_i$, respectively. Then, we quantify the \textit{Removal Error} ($\epsilon^{rem}$) as the proportion of ground-truth minutiae lost during generation. Similarly, the \textit{Addition Error} ($\epsilon^{add}$) is quantified as the proportion of minutiae in the synthetic image that are hallucinated. Thus,

\begin{equation*}
\epsilon^{rem}_i = \frac{\beta_i}{\alpha_i+\beta_i} ~~\text{and}~~ \epsilon^{add}_i = \frac{\gamma_i}{\alpha_i+\gamma_i}.       
\end{equation*}

\begin{figure*}[h]
\centering
\includegraphics[width=0.8\linewidth]{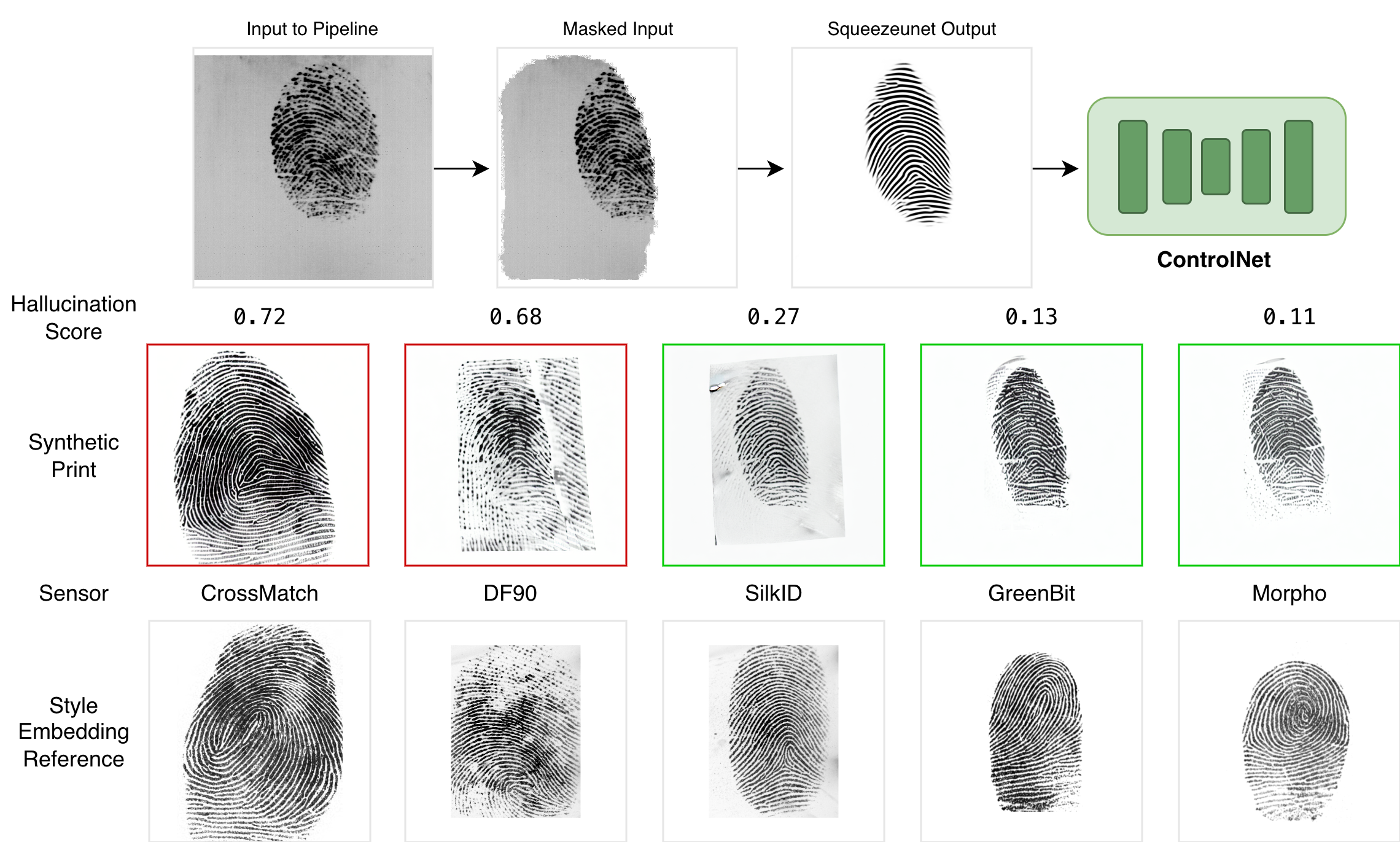} 
\caption{Qualitative comparison of global hallucination across different sensor styles. The \textcolor{red}{red} and \textcolor{lime}{green} boundaries indicate high and low hallucination scores $(1-\text{IoU})$, respectively. For all the synthetic prints (middle row), the ridge pattern guidance is the same (top row) but the style guidance is different (bottom row).} 
\label{fig:diff_sensors_global_hal}
\end{figure*}
\vspace{1cm}

\begin{table}[h!]
\caption{Local errors (addition and removal of minutiae points) introduced by the generation pipeline. }
\begin{center}
\resizebox{0.6\columnwidth}{!}{%
\begin{tabular}{@{}lccc@{}}
\toprule
\textbf{Local} & \multicolumn{2}{c}{\textbf{Input Quality}} & \\
\cmidrule(lr){2-3}
\textbf{Error Type} & \textbf{\textit{High}}& \textbf{\textit{Low}} & \textbf{Total} \\
\midrule
Removal (\%) & 11.05 & 13.83 & 12.42\\
Addition (\%) & 10.98 & 31.67 & 21.45 \\
\bottomrule
\end{tabular}}
\label{tab:manual_annotation}
\end{center}
\vspace{-0.3cm}
\end{table}

\begin{figure*}[h!]
\centering
\includegraphics[width=0.7\linewidth]{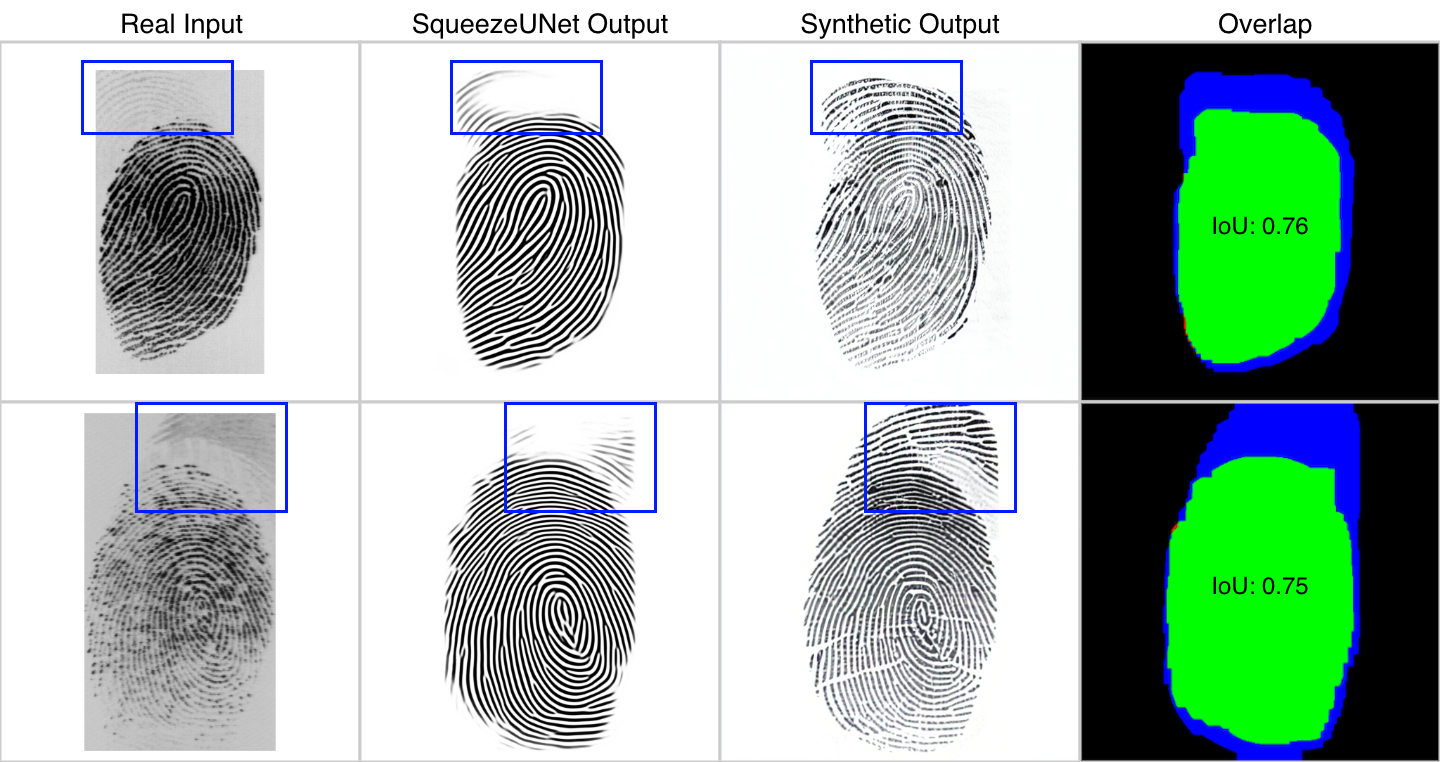} 
\caption{Examples of extrapolation due to enhancement errors.}
\label{fig:hal_example_2}
\end{figure*}

\noindent The average removal and addition errors over the dataset are considered as metrics to quantify the local errors introduced during generation. The results in \cref{tab:manual_annotation} demonstrate that in general, the generation pipeline does not preserve 100\% of the ground-truth minutiae. On average, $87.58\%$ of the minutiae are preserved with an addition error of $21.45\%$. In particular, there is a relationship between the quality of the exemplar print and the local errors made by the generator. If the quality of the exemplar print is high, the addition error is only $10.98\%$ compared to a $31.67\%$ addition error for low quality exemplars. While the removal error is not very sensitive to the input quality, a low quality input still causes more removal error than a high quality input.  We attribute these errors to following aspects of the pipeline: (1) SqueezeUNet enhancement: The SqueezeUNet in the pipeline is responsible for denoising and extracting the ridge skeleton from the exemplar print. In the case of low-quality exemplars, SqueezeUNet can over-smooth faint ridges, leading to removal of minutiae as shown in \cref{fig:example_manual_annotation}b. In contrast, new minutiae may be added due to incorrect enhancement of low-quality exemplars as shown in \cref{fig:example_manual_annotation}c. Both cases could occur as shown in \cref{fig:example_manual_annotation}d, leading to high removal and addition of minutiae. An example of 100\% correspondence (perfect identity preservation) is shown in \cref{fig:example_manual_annotation}a. (2) Style-induced errors: The injected style guidance also plays a minor role in inducing local errors. When the target style contains high-frequency texture, the diffusion model may hallucinate minutiae where none exist.

\begin{figure}[h]
\centering
\includegraphics[width=\linewidth]{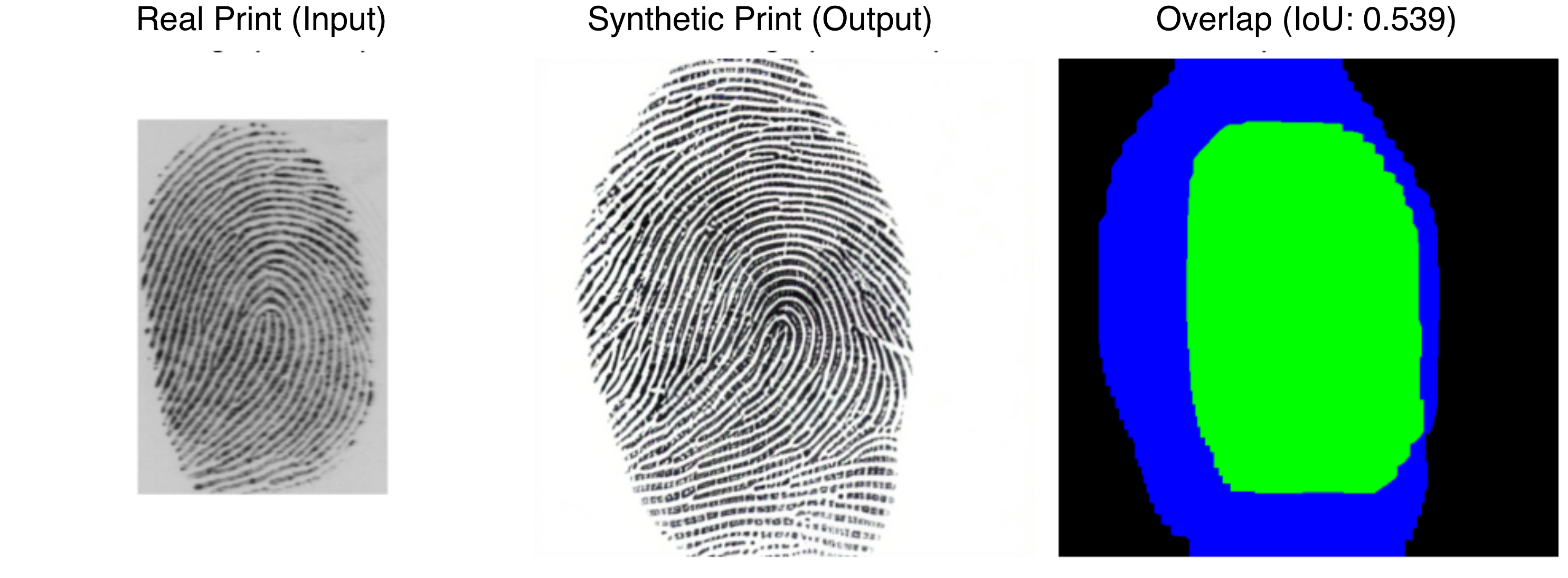} 
\caption{Example of global ridge pattern hallucination. \textcolor{latentgreen}{Green} corresponds to the overlapped region between expected $\mathbf{B}^{exp}_i$ and synthetic print $\mathbf{B}^{gen}_i$ masks. \textcolor{latentblue}{Blue} shows the hallucinated region.}
\label{fig:example_global_hal}
\end{figure}

\noindent \textbf{Global Errors}:
While local minutiae errors are introduced within the expected fingerprint region (foreground mask), the diffusion model also extrapolates (hallucinates) ridge patterns in the synthetic print beyond the expected foreground mask. Such an extrapolation can be considered as global error introduced by the generative process. Diffusion models are inherently designed to synthesize continuous and coherent textures. When presented with a mask guidance, the model often fails to distinguish between missing ridge patterns within the foreground mask that needs inpainting and a valid background. Consequently, the generative process extrapolates the ridge pattern into the background space, effectively increasing the fingerprint's surface area. To quantify this extrapolation, we measure the Intersection over Union (IoU) between the expected mask $\mathbf{B}^{exp}_i$ and the mask extracted from the synthetic print  $\mathbf{B}^{gen}_i$. 

\begin{table}[h]
\caption{Global hallucination error rates for different sensor styles.}
\centering
\resizebox{0.7\columnwidth}{!}{%
    
    \begin{tabular}{@{}lc@{}}
    \toprule
    \textbf{Sensor} & \textbf{Hallucination Error Rate (\%)} \\
    \midrule
    CrossMatch & $30.04 \pm 1.025$ \\
    Futronic &  $13.00 \pm 0.575$ \\
     DF90 &  $12.30 \pm 0.851$ \\
     GreenBit & $7.30 \pm 0.299$ \\
     Morpho & $7.19 \pm 0.256$ \\
     SilkID &  $6.78 \pm 0.448$\\
     \bottomrule
    \end{tabular}
    }
\label{tab:global_hallucination_results}
\end{table}

\noindent The \textit{Global Hallucination Error Rate (\%)} is defined as the fraction of synthetic prints in the dataset having $\text{IoU}<0.8$. An example of such an error is shown in \cref{fig:example_global_hal}.  Our analysis of global hallucination (\cref{tab:global_hallucination_results} and \cref{fig:diff_sensors_global_hal}) reveals that the size mismatch between the expected foreground mask and the target style embedding is a primary driver of global error. Global hallucination errors are more common for some sensor styles like CrossMatch ($30.04\%$ in our experiments) compared to other sensors for the same acquisition type (slap). This occurs because the CrossMatch sensor typically captures a larger fingerprint area shown in \cref{fig:diff_sensors_global_hal}. When the pipeline receives a small expected mask but is conditioned on a style with a much larger fingerprint foreground region, the diffusion model attempts to mimic the target style by extrapolating ridges into the regions outside the expected mask. This explains why different sensor styles cause different global error rates despite receiving identical ridge guidance. Additionally, the SqueezeUNet backbone can occasionally fail to distinguish between valid ridges and background noise or blurred ridges in the exemplar input image. The ControlNet reinforces these false background ridges, resulting in a synthesized fingerprint that could be larger than expected, as shown in \cref{fig:hal_example_2}.

\section{Conclusion}
This work explored the intra-finger variability of synthetic fingerprints generated by a SOTA diffusion model called GenPrint. While our research applies to all types of fingerprints (rolled, plain, and latent prints), we primarily focus on latent prints because of the scarcity of public domain (latent, rolled) paired databases. We first increase the diversity of synthetic latent fingerprints generated by GenPrint, by covering 40 distinct surfaces where latent impressions could be formed and 15 different processing techniques to effectively “lift” the latents. Next, we investigate the identity consistency of synthetic prints and show that diffusion models are prone to hallucinations. To mitigate this problem, we recommend explicitly incorporating ridge structure and minutiae constraints into the loss function used to train the diffusion model.


\section*{Acknowledgments}

This work was supported by the National Institute of Standards and Technology 60NANB25D116-0.

{   
    \small
    \bibliographystyle{ieeenat_fullname}
    \bibliography{main}
}

\end{document}